\documentclass[lettersize,journal]{IEEEtran}
\usepackage{amsmath,amsfonts}
\usepackage{algorithmic}
\usepackage{algorithm}
\usepackage{multirow} 
\usepackage{array}
\usepackage[caption=false,font=normalsize,labelfont=sf,textfont=sf]{subfig}
\usepackage{textcomp}
\usepackage{stfloats}
\usepackage{url}
\usepackage{verbatim}
\usepackage{graphicx}
\usepackage{cite}
\usepackage{hyperref}
\usepackage{booktabs} 
\usepackage{multirow} 
\usepackage{siunitx} 
\usepackage{amssymb} 
\hypersetup{colorlinks=true,
            linkcolor=blue,
            anchorcolor=blue,
            citecolor=blue}
\usepackage{cleveref}
\usepackage{booktabs}
\begin{document}

\title{Raformer: Redundancy-Aware Transformer \\ for Video Wire Inpainting}

\author{Zhong~Ji, \textit{Senior Member, IEEE},
        Yimu~Su,
        Yan~Zhang,
        Jiacheng~Hou,
        \\
        Yanwei~Pang, \textit{Senior Member, IEEE},
        Jungong Han, \textit{Senior Member, IEEE}
        
\thanks{This work was supported by the National Key Research and Development Program of China (Grant No. 2022ZD0160403), and the National Natural Science Foundation of China (NSFC) under Grant 62176178 (Corresponding author: Yan Zhang).

Zhong Ji and Yanwei Pang are with the School of Electrical and Information Engineering, Tianjin Key Laboratory of Brain-Inspired Intelligence Technology, Tianjin University, Tianjin 300072, China, and also with the Shanghai Artificial Intelligence Laboratory, Shanghai 200232, China (e-mail: jizhong@tju.edu.cn; pyw@tju.edu.cn).

Yimu Su, Yan Zhang, and Jiacheng Hou are with the School of Electrical and Information Engineering, Tianjin Key Laboratory of Brain-Inspired Intelligence Technology, Tianjin University, Tianjin 300072, China (e-mail: suyimu@tju.edu.cn; yzhang1995@tju.edu.cn; hjc@tju.edu.cn ).

Jungong Han is Chair Professor in Computer Vision at the Department of Computer Science, the University of Sheffield (e-mail:jungonghan77@gmail.com).
}

\markboth{Journal of \LaTeX\ Class Files,~Vol.~14, No.~8, August~2024}%
{Shell \MakeLowercase{\textit{et al.}}: A Sample Article Using IEEEtran.cls for IEEE Journals}
}

\maketitle
\begin{abstract}
Video Wire Inpainting (VWI) is a prominent application in video inpainting, aimed at flawlessly removing wires in films or TV series, offering significant time and labor savings compared to manual frame-by-frame removal. However, wire removal poses greater challenges due to the wires being longer and slimmer than objects typically targeted in general video inpainting tasks, and often intersecting with people and background objects irregularly, which adds complexity to the inpainting process.
Recognizing the limitations posed by existing video wire datasets, which are characterized by their small size, poor quality, and limited variety of scenes, we introduce a new VWI dataset with a novel mask generation strategy, namely Wire Removal Video Dataset 2 (WRV2) and Pseudo Wire-Shaped (PWS) Masks. WRV2 dataset comprises over 4,000 videos with an average length of 80 frames, designed to facilitate the development and efficacy of inpainting models. Building upon this, our research proposes the Redundancy-Aware Transformer (Raformer) method that addresses the unique challenges of wire removal in video inpainting. Unlike conventional approaches that indiscriminately process all frame patches, Raformer employs a novel strategy to selectively bypass redundant parts, such as static background segments devoid of valuable information for inpainting. At the core of Raformer is the Redundancy-Aware Attention (RAA) module, which isolates and accentuates essential content through a coarse-grained, window-based attention mechanism. This is complemented by a Soft Feature Alignment (SFA) module, which refines these features and achieves end-to-end feature alignment. Extensive experiments on both the traditional video inpainting datasets and our proposed WRV2 dataset demonstrate that Raformer outperforms other state-of-the-art methods. Our codes and the WRV2 dataset will be made available at: \href{https://github.com/Suyimu/WRV2}{https://github.com/Suyimu/WRV2}.
\end{abstract}


\section{Introduction}
\IEEEPARstart{V}{ideo} inpainting aims at filling in videos with holes by padding temporally-consistent content, where the holes represent occlusion or unwanted objects\cite{TIPyang2023deep,IIVI2021,TIPjia2022non,TIPpatwardhan2007video,ECCVerror,STTN,Pronpainter_2023_ICCV,TIPebdelli2015video}. The fundamental goal of video inpainting is to enhance the visual coherence and continuity of inpainted video streams, especially in scene editing. By effectively erasing or modifying superfluous elements, it contributes significantly to the refinement of visual storytelling, ensuring the seamless integration of visual elements in diverse contexts. 

Different from traditional object-removal video inpainting methods\cite{STTN,IIVI2021}, Video Wire Inpainting (VWI) specifically targets the elimination of wires, which are commonly employed in action sequences for special effects. 
This focus on wires is particularly relevant in film and television production, where wire removal\cite{collis2004wire} is a key technique employed to facilitate special effects and action sequences cost-effectively during shooting. 
Besides, wire removal is always crucial in post-production to maintain visual integrity. 
\begin{figure}[!t]
\centering
\includegraphics[width=1\columnwidth]{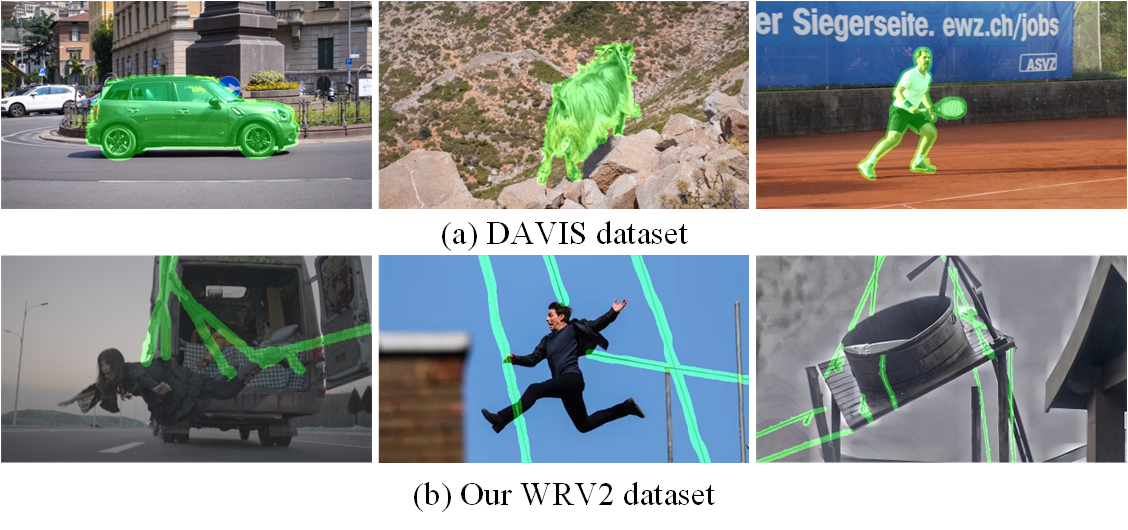}
\caption{Comparative Illustration: (a) DAVIS Dataset\cite{DAVIS} vs. (b) Our WRV2 Dataset. The upper panel displays typical object masks from the DAVIS dataset, while the lower panel highlights our dataset with a specific focus on wire masks, necessitating removal in film and television post-production.}
\label{WRV_DAVIS_masks}
\end{figure}
This meticulous process significantly extends the duration of post-production, reflecting the intricate and precision-driven nature of professional video editing.
Traditionally, wire removal has been a labor-intensive and time-consuming task in film production. It involves professionals meticulously scanning each video frame, manually selecting wire areas, and employing graphic editing tools like ``content-aware fill'' or ``clone stamp''\cite{oh2007image} to blend the wires with the surrounding background or texture. This conventional method not only requires extensive manual labor but also runs the risk of inconsistency, potentially leading to uneven or unnatural outcomes in the final video. 

Thus, achieving automatic wire removal has garnered increased attention \cite{ji2022g2lp,Chiu_2023_CVPR}. However, the unique characteristics of this task make it more challenging than traditional object-removal inpainting tasks, as illustrated in Fig. \ref{WRV_DAVIS_masks}.
For example, the wires in our context are typically long and slender, often stretching across the entire image while being only a few pixels wide. 
This nature of wire shape masks makes it challenging to achieve seamlessly integrated inpainting areas, especially when they overlap with multiple objects, thereby complicating the removal process more than with regular object masks.

Faced with the above characteristics, the significance of a comprehensive benchmark dataset becomes paramount in advancing the field of Video Wire Inpainting (VWI). Access to suitable raw footage that accurately represents the specific requirements of wire removal in a film production context is increasingly challenging. Most videos available online have already undergone post-production processing, making it difficult to find unedited sequences. Despite the availability of video inpainting datasets like YouTube-VOS\cite{xu2018youtube} and DAVIS\cite{DAVIS}, they fall short in capturing the unique visual challenges posed by unedited film scenes, such as actors performing aerial stunts or other actions facilitated by wire techniques. Our previous work, G2LP \cite{ji2022g2lp}, embarked on an initial exploration in VWI by releasing the Wire Removal Video Dataset (WRV) suitable for evaluation and employing a global-to-local transformer approach to tackle VWI challenges. However, due to its limited scale and less accurate annotations, the WRV dataset falls short in training high-quality models capable of effectively solving the VWI task.

\begin{figure}[!t]
\centering
\includegraphics[width=1\columnwidth]{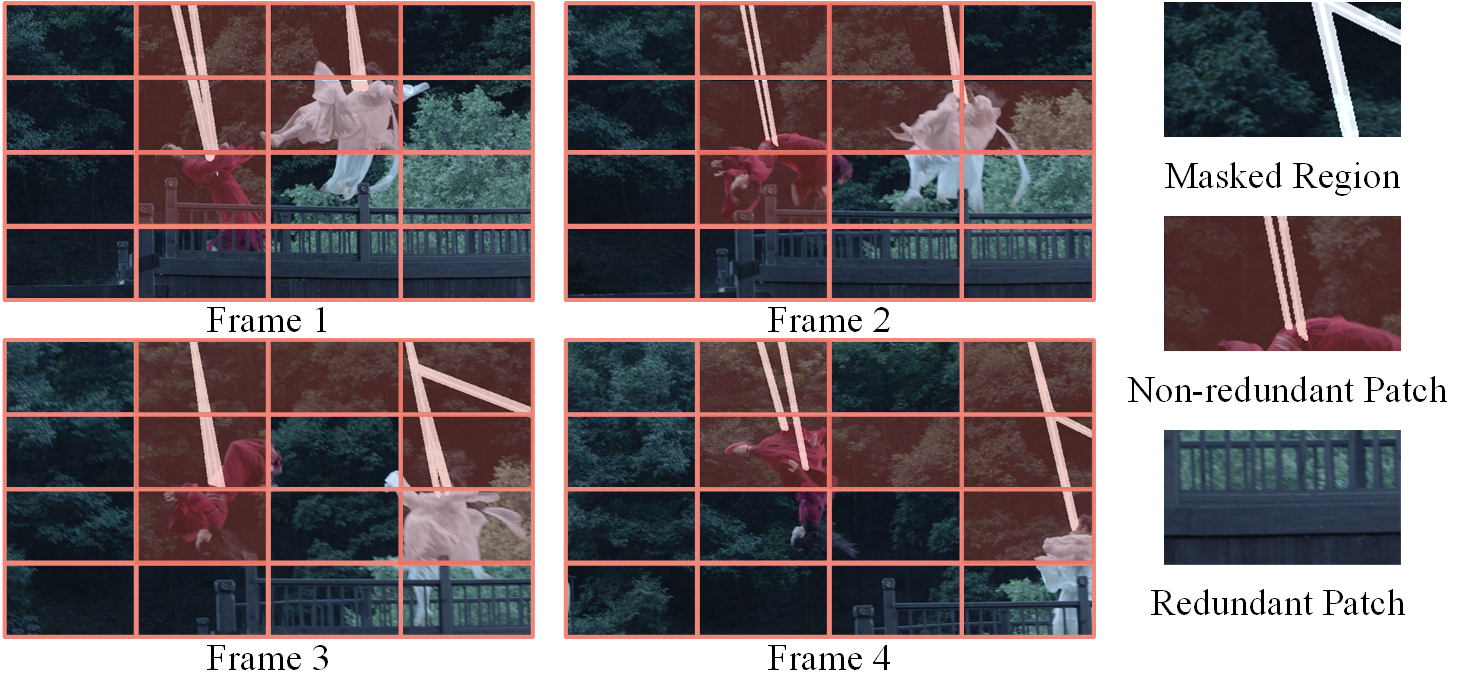}
\caption{The main idea of Raformer, 
where four frames are illustrated. The masked region includes the wires that require completion, the non-redundant patch contains valuable information for achieving video wire inpainting, and the redundant patch should be eliminated as it is unnecessary. It is worth noting that non-redundant patches are not fixedly limited to those that contain wires.}
\label{house_masks}
\end{figure}

Another significant challenge in advancing VWI is that advanced video inpainting approaches\cite{wu2023semi,zhang2022inertia,wu2023deep,li2023short} often face challenges and even fail to address the VWI task. The reason stems from traditional inpainting frameworks being geared towards removing occlusions or objects that occupy larger, more regularly shaped areas in frames, as opposed to slimmer and longer shaped elements like wires as illustrated in Fig. \ref{WRV_DAVIS_masks}. Methodologically, these approaches relies on borrowing patches from other frames to replenish missing content, where each patch is assigned with equal importance. Nevertheless, few research efforts have been specifically dedicated to addressing these issues. In this work, we argue that different patches hold varying levels of importance to the inpainting process, where certain patches provide valuable information and others are redundant, as shown in Fig. \ref{house_masks}. Specifically, in a scene where two individuals are engaged in a roll maneuver facilitated by wire techniques, certain patches, like the one in the lower right corner of frame 2, neither contribute to the inpainting process nor offer relative positional references. Conversely, the marked "Non-redundant patches" furnish the majority of the reference information for the missing areas. This realization leads us to strategically identify and prioritize key areas within a video frame, focusing our attention on valuable information and enabling the model to concentrate on patches that significantly affect the quality and accuracy of the inpainted video, thereby yielding precise and contextual results.

To this end, we introduce Wire Removal Video Dataset 2 (WRV2), a more comprehensive wire-removal dataset compared to its predecessor WRV\cite{ji2022g2lp}, along with a novel algorithm for generating Pseudo Wire-Shaped (PWS) Masks, designed to mirror real-world wire scenarios for both training and evaluation. Additionally, we present the Redundancy-Aware Transformer (Raformer), a method specifically designed to address the intricate demands of wire removal in film production, i.e., the VWI task. Methodologically, Raformer integrates two innovative modules: the Redundancy-Aware Attention (RAA) Module and the Soft Feature Alignment (SFA) Module.

Our main contributions are summarized as follows:
\begin{itemize}
\item {We present the WRV2, a comprehensive dataset of 4,232 unedited video sequences tailored for VWI. This dataset includes wire-specific annotations and a novel algorithm for generating PWS masks, making it a pivotal resource for film and industry-related inpainting tasks.}
\item{We introduce the Raformer, a novel method specifically designed for the VWI task. Raformer innovatively addresses the challenges of wire removal by focusing on the valuable patches and  eliminating redundant information, significantly enhancing the accuracy of the wire removal process.}
\item{We integrate two innovative modules in Raformer: the RAA Module and the SFA Module. The RAA module boosts the inpainting process by employing window partitioning and spatio-temporal attention to eliminate redundancy at a coarse level in each video frame, while the SFA module restores and aligns the features, ensuring seamless integration and natural aesthetics in the inpainted videos.}

\end{itemize}
\section{Related Work}
\subsection{Image Inpainting}
Image inpainting\cite{xu2010image,TIPimageghorai2019multiple,yu2019free,liu2020rethinking,TIPimagekim2023progressive,TIPxue2017depth}, essential in digital image processing, fills missing or damaged areas in images. Employed in restoring photographs and artworks, it now plays a key role in modern digital editing. Image inpainting mends gaps, ensuring the original image's consistency. Relatedly, video inpainting builds upon these principles, but adds the complexity of maintaining temporal coherence across frames. 

Recent advancements in image inpainting focus mainly on GAN-based\cite{yu2018generative,TIPimagedeng2023context}, transformer-based\cite{li2022mat,tcsvtimageli2023transformer}, and diffusion-based models\cite{rombach2022high,lugmayr2022repaint}. GAN-based models are known for creating realistic images, transformer-based models excel in handling complex tasks with their ability to capture contextual information, and diffusion-based models are notable for generating highly realistic textures. Each approach has its strengths and challenges, with GANs sometimes producing artifacts, transformers requiring significant computational resources, and diffusion models being slower in processing.

A notable contribution to wire removal is the work of Chiu et al. \cite{Chiu_2023_CVPR}, which tackles the unique challenges of thin, lengthy, and sparse wires in high-resolution images with a two-stage method that efficiently combines global and local context for precise wire segmentation. While this represents a significant advancement in image inpainting for wire removal, it's crucial to acknowledge that the VWI task, with its need for temporal consistency, presents distinct challenges that extend beyond the scope of single-image processing.

\subsection{Video Inpainting} 
Video inpainting\cite{fuseformer,TIPhou2024mcd, lee2019copy,kim2019recurrent,li2020short}, extended from image inpainting, tackles the additional challenge of maintaining temporal consistency and motion coherence in video sequences. This evolution from static image restoration to dynamic video processing necessitates seamless continuity across entire sequences, adding complexity to the task. 

We broadly categorize the recent advancements in video inpainting into four main groups: 3D Convolution-based, Internal Learning-based, Flow-Guided Propagation-based, and Video Transformer-based approaches. 3D Convolution, employing 3D CNNs\cite{Free-form,wang2019video}, processes spatiotemporal data effectively but often struggles with limited receptive fields and frame misalignment. Parallel to this, Internal Learning methods focus on leveraging the inherent content to encode and memorize appearance and motion for inpainting\cite{IIVI2021,ren2022dlformer}. However, these approaches require individual training for each video, posing challenges to its practicality and scalability. Following these methods, Flow-Guided Propagation has also emerged as a key technique in video inpainting. It utilizes optical flow\cite{E2FGVI,chang2019learnable,li2020short,zou2021progressive} to align frames and maintain temporal coherence. 

Shifting the focus to Video Transformers, a pioneering work in this field is STTN \cite{STTN}, which integrates ViT \cite{dosovitskiy2020image} into video inpainting, effectively addressing both spatial and temporal aspects. Liu et al. \cite{fuseformer} further evolved this approach, enhancing transformer-based methods with fine-grained information to better replicate the dynamics in video sequences. The integration of optical flow with transformer takes a great leap in this field, improving alignment and consistency in videos, exemplified by \cite{E2FGVI,zhang2022FGT}. Moreover, Zhou et al. \cite{Pronpainter_2023_ICCV} combined dual-domain propagation with a mask-guided sparse transformer and effectively tackled spatial misalignment and computational challenges in processing dense video data.

Despite the significant advancements achieved by the above methods, the methods of recognizing and addressing the redundancy issues inherent in video transformer remain largely manual and mechanized \cite{Pronpainter_2023_ICCV,E2FGVI,SwinVI}. By contrast, 
inspired by the recent developments of transformer \cite{tcsvttang2023datfuse,rao2021dynamicvit,yin2022vit,kong2021spvit}, our method considers the significance of different sections and enables autonomously eliminate the non-redundant information at a coarse granularity of features. 
Most closely related to our work is \cite{ji2022g2lp}, which employs a global-to-local transformer and introduces a WRV dataset with 150 sequences. Differently, in this work, we focus on redundancy elimination and introduce a new VWI dataset to mitigate the weaknesses of the previous version.

\begin{table}[t]
    \centering
    \caption{Summary of the Datasets}
    \label{tab: dataset}
    \begin{tabular}{cccc}
    \toprule
    Dataset & Videos & Average Frames & Annotation \\
    \midrule
    DAVIS\cite{DAVIS} & 150 & ~70& Object annotation\\
    YouTube-VOS\cite{xu2018youtube} & 4453 & ~135& Object  annotation\\
    WRV\cite{ji2022g2lp} & 150 & ~40& Wire annotation\\
    WRV2 & 4232& ~80& Wire annotation\\
    \bottomrule
    \end{tabular}
\end{table}

\section{Wire Removal Video Dataset 2}
\subsection{The Characteristics of WRV2 Dataset}

\begin{figure}[!b]
    \centering
    \includegraphics[width=1\linewidth]{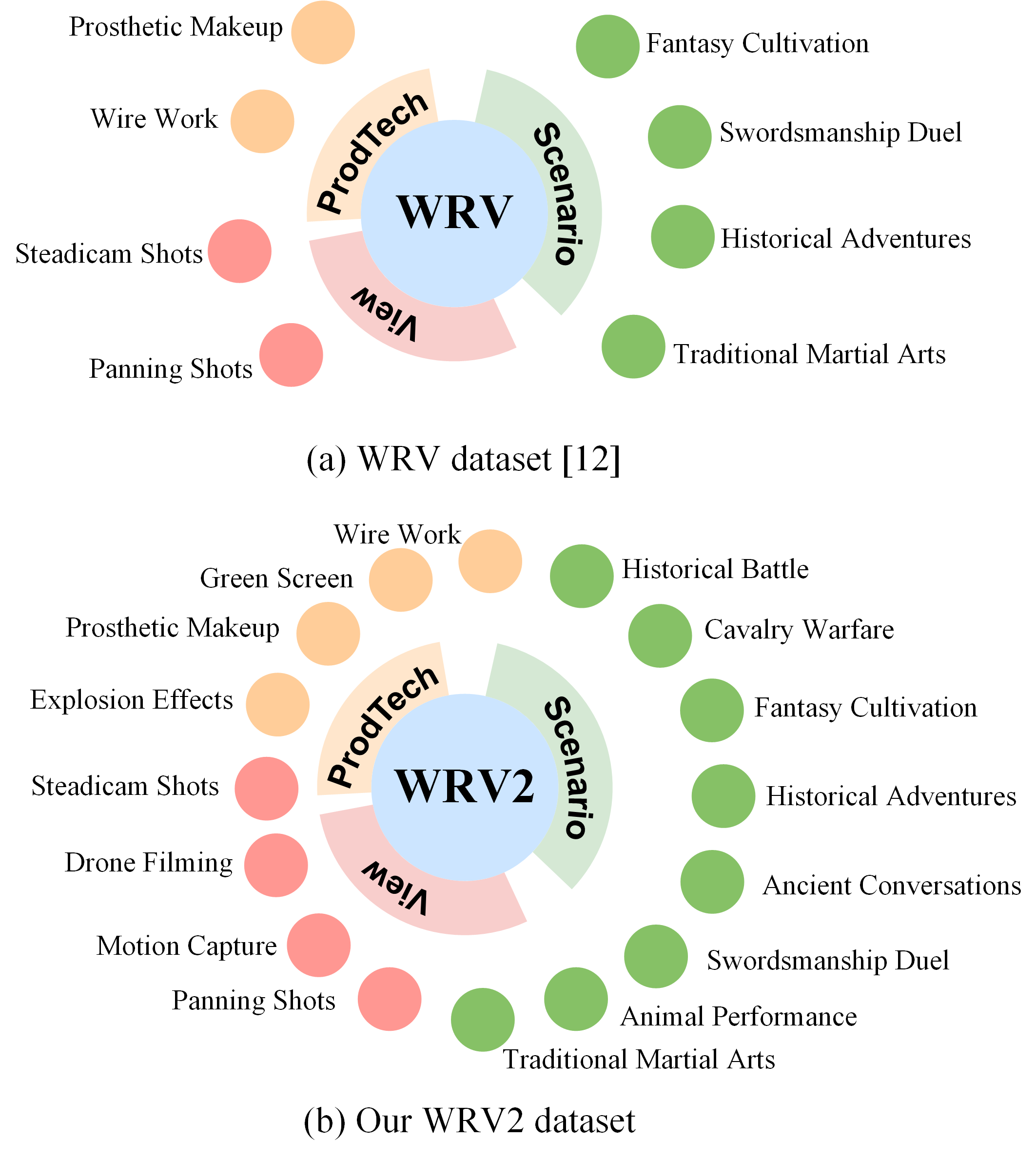}
    \caption{Overview of both the (a) WRV and (b) WRV2 datasets from three perspectives:  View, Scenario, and ProdTech (an abbreviation for Production Technique). }
    \label{fig_dataset2}
\end{figure}

We summarize the existing video inpainting datasets and the VWI datasets in Table \ref{tab: dataset}, including DAVIS\cite{DAVIS}, YouTube-VOS\cite{xu2018youtube}, WRV\cite{ji2022g2lp}, and our released WRV2. The first two datasets are traditional video inpainting datasets, which fall short of fully meeting the demands of this specialized task. 
Specifically, the DAVIS dataset provides 150 video sequences primarily for segmentation, which is proposed and primarily employed for video segmentation tasks. 
Thus, it lacks the diversity and scale needed for the VWI task such as a variety of scenarios and wire masks. Similarly, although the YouTube-VOS dataset possesses over 4,000 videos, it still fails to meet the requirement of VWI task, particularly in terms of scenes with wire occlusions and detailed wire mask annotations. 
These gaps underscore the urgent need for a dataset tailored for VWI. 

\begin{figure*}[!t]
\centering
\includegraphics[width=0.95\textwidth]{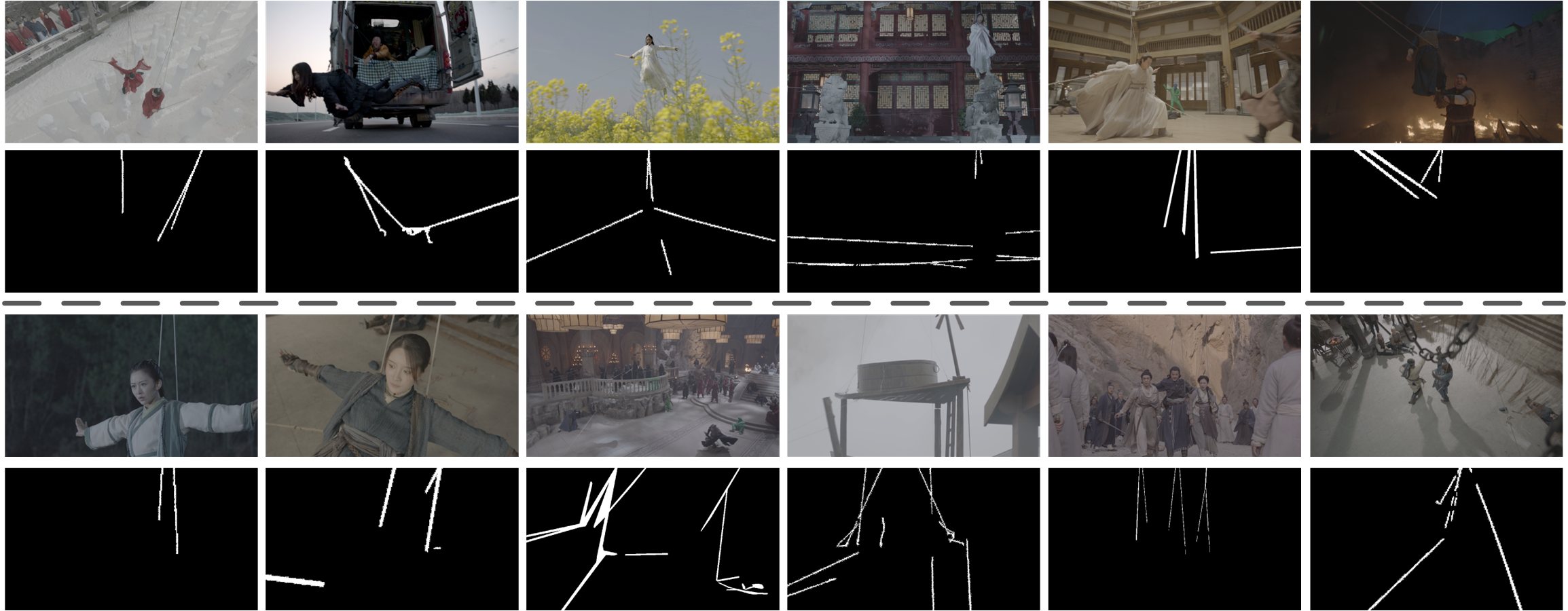}
\caption{Description of the dataset image. The first and third rows feature scene images, while the second and fourth rows present their corresponding mask annotations.}
\label{fig_dataset}
\end{figure*}

Notably, there is already a publicly released VWI dataset, which is our previous work, i.e., WRV \cite{ji2022g2lp}. In detail, the WRV dataset is comprised of  150 video clips from movies and TV series, which provides a realistic context, particularly in scenes with slender and irregular areas. 
However, its limited size restricts its applicability for extensive training purposes, as it fails to provide a comprehensive range of scenarios necessary for robust algorithm development.
We argue that a specialized VWI dataset, crucial for effective model training, should encompass a wide range of scenarios with diverse backgrounds, lighting, and realistic complexities, along with accurate wire masks and targeted annotations. 

To this end, we develop an expansive and meticulously curated dataset, which is abbreviated as WRV2. Fig. \ref{fig_dataset2} clearly shows the detailed information of our WRV2 dataset and its significant improvement compared with the previous WRV dataset\cite{ji2022g2lp}.
Specifcally, WRV2 dataset is compiled from a collection of raw, unedited video sequences—uncommon in visual effects.
Comprising 4,232 video sequences, WRV2 provides a rich and diverse landscape for wire inpainting tasks. 
These sequences originate from the same filming contexts as the wire effects but feature a variety of other special effects. This includes, but is not limited to, explosive effects, ancient makeup masks, and diverse scenarios that are contextually aligned with the wire scenes. 
Fig. \ref{fig_dataset} showcases selected frames and masks from the WRV2 dataset, highlighting its diversity and detail.
Structurally, WRV2 is divided into training and testing sets, with 3,762 sequences allocated for training purposes and 470 designated for testing. This testing subset includes 220 real-life wire removal scenarios with authentic wire masks (manually annotated) and 250 sequences with artificially generated masks, a critical element for assessing the robustness of inpainting algorithms. It is worth noting that the 470 sequences dedicated to testing in WRV2 include all content from WRV. The methodology for generating these PWS masks is detailed in subsequent sections of the paper. 
\subsection{Pseudo Wire-Shaped Masks Generation}
\begin{algorithm}[t]
\caption{Pseudo Wire-Shaped Masks Generation}\label{alg:alg1}
\begin{algorithmic}[1]
\REQUIRE $\mathbf{num, shape,len},X$
\ENSURE Image from canvas $X$, Wire-shaped Video Mask list

\STATE \textbf{Function} {\textsc{CREATE WIRE MASK}}$(\mathbf{num,shape})$
\STATE \hspace{0.5cm} Initialize an empty canvas $X$
\STATE \hspace{0.5cm} Define the range for wire width and length
\FOR{each  $\mathbf{wire}$ in $\mathbf{num}$}
\STATE \hspace{0.5cm} Generate a wire randomly within $\mathbf{shape}$
\STATE \hspace{0.5cm} Apply a random affine transformation
\STATE \hspace{0.5cm} Perform random translation and cropping
\STATE \hspace{0.5cm} Add the generated wire to the canvas $X$
\ENDFOR
\STATE \hspace{0.5cm} Resize $X$ to the desired size
\STATE \hspace{0.5cm} Employ a kernel function to dilate $X$ randomly within $\mathbf{num}$ times
\STATE \hspace{0.5cm} Detect the boundaries of the wire mask and crop accordingly
\RETURN Image from canvas $X$
\STATE \textbf{Function} {\textsc{CREATE VIDEO MASK}}$(\mathbf{X,len})$
\STATE \hspace{0.5cm} Initialize a list of images with length $\mathbf{len}$
\FOR{$i = 1$ to $\mathbf{len}$}
\STATE \hspace{0.5cm} Randomly initialize a position
\STATE \hspace{0.5cm} Randomly assign a movement amount
\STATE \hspace{0.5cm} Apply the position and movement to $X$
\STATE \hspace{0.5cm} Add $X$ to the images list
\ENDFOR
\RETURN Wire-shaped Video Mask list
\end{algorithmic}
\end{algorithm}
Traditional video inpainting methods, exemplified by \cite{E2FGVI,fuseformer, STTN}, have predominantly employed Polygonal Pseudo (PP) masks\cite{STTN} that manifest as irregular polygons. This approach is limited in the real scene, especially for wire removal tasks where these masks tend to obscure distinct entities and perform subpar, as demonstrated in Section \ref{sec:PWS}. Recognizing the need for better-suited solutions in such scenarios, we propose ``Pseudo Wire-Shaped (PWS) masks" as a superior alternative specifically designed for these tasks. This novel mask generation technique aims to improve performance by fitting more accurately to the shapes needing removal, thereby optimizing the training and evaluation process in VWI tasks.

Algorithm \ref{alg:alg1} presents the pseudocode of generating PWS masks, where ``num'' denotes the number of wires, ``shape'' specifies the constraints on the wire's length and width,  ``len'' represents the duration of the video, and ``X'' is initialized as an empty canvas. The process involves drawing each wire on the canvas as per the given specifications and randomly applying affine transformations to simulate real-world irregularities. Each wire is then randomly placed to vary in appearance and orientation across the canvas. After all wires are added, the canvas is resized and processed using a kernel function to finalize the masks. A sequence of video masks is also created by dynamically changing positions and applying movements to the canvas, simulating natural motion, which helps in training models to handle wire removal effectively in various video contexts.

\begin{figure}[!t]
\centering
\includegraphics[width=0.95\columnwidth]{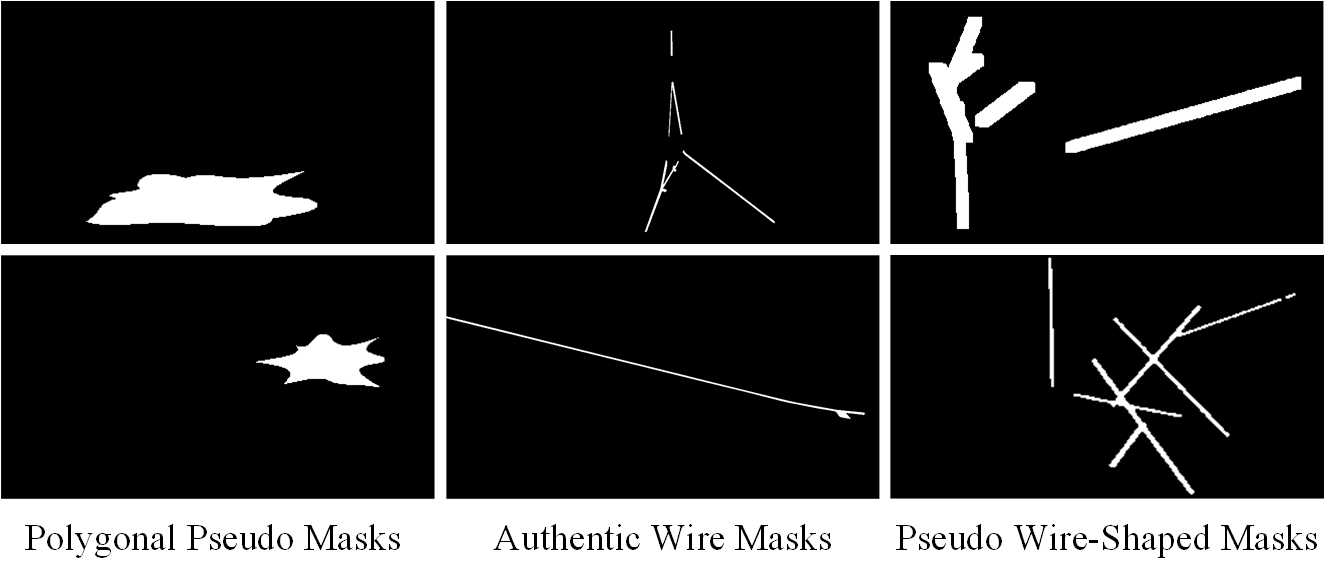}
\caption{Illustration of different types of masks, including Polygonal Pseudo Masks, Authentic Wire Masks, and our proposed Pseudo Wire-Shaped Masks, with the latter showcasing masks produced under two different parameter settings.}
\label{fig_masks}
\end{figure}

As illustrated in Fig. \ref{fig_masks}, our PWS masks offer a dynamic and adjustable solution unlike the static traditional PP masks, better replicating the typical wire occlusions found in film and television. These masks can be modified in both length and width, enabling realistic simulations of various wire scenarios that mimic real wire complexities across the image. 
Unlike PP masks covering large areas with irregular shapes and lacking the detail needed for nuanced tasks like wire removal, PWS masks feature slim, elongated lines better suited for such specific challenges. Training with PWS masks is particularly advantageous for the VWI task as they more effectively replicate real-world wire removal challenges, ensuring a more precise and efficient inpainting process. For a detailed quantitative analysis, please refer to Section \ref{sec:PWS}.

\section{Approach}
\begin{figure*}[t]
\centering
\includegraphics[width=0.98\textwidth]{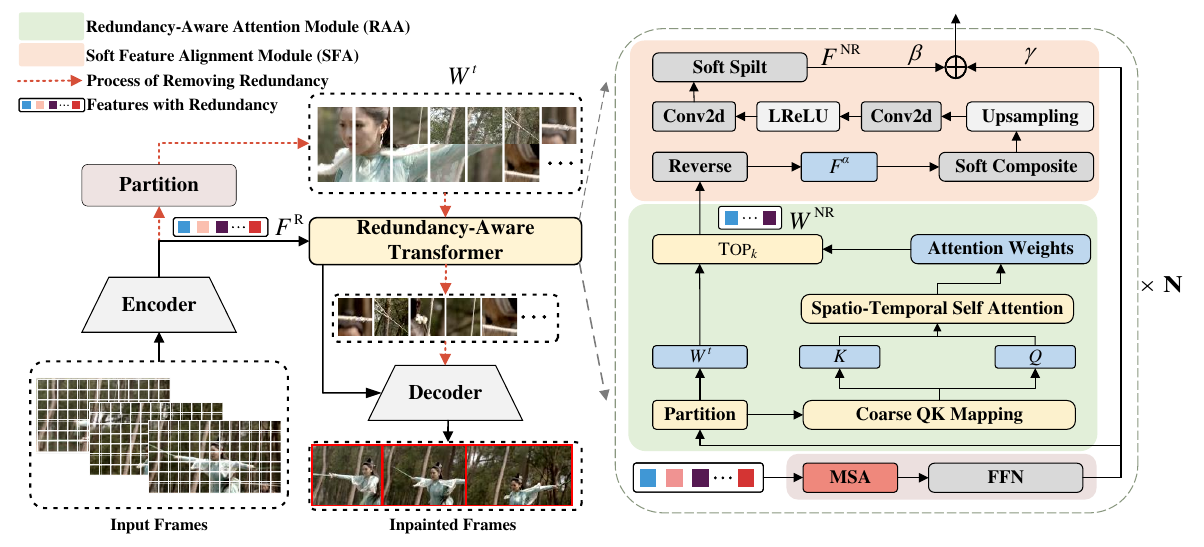}
\caption{Illustration of our proposed Redundancy-Aware Transformer (Raformer) and the workflow of Video Wire Inpainting. Note that the operations of Multihead Self-Attention (MSA) and Feedforward Neural Network (FFN) are the same as\cite{Pronpainter_2023_ICCV}. Additionally, the red dashed lines indicate a visualization flow and are not part of the actual implementation.
}
\label{fig_model}
\end{figure*}
In the realm of VWI, given the input data \(\textbf{X} = \left\{ X_t \in \mathbb{R}^{H \times W \times 3} \, \big| \, \forall t \in \{1, 2, \ldots, T\} \right\} \) and its corresponding binary mask \(\textbf{M} = \left\{ M_t \in \mathbb{R}^{H \times W \times 1} \, \big| \, \forall t \in \{1, 2, \ldots, T\} \right\} \), where the binary values of \textbf{M} encode the presence or absence of valid information, the fundamental objective of VWI is crystallized in the synthesis of a regenerated video \( \textbf{Y} = \left\{ Y_t \in \mathbb{R}^{H \times W \times 3} \, \big| \, \forall t \in \{1, 2, \ldots, T\} \right\} \).  This synthesis entails the harmonious fusion of \textbf{X} under the selective guidance of \textbf{M}, encapsulating the intricate process of inferring and completing missing visual content. In this setup, \(H\)\ and \(W\)\ denote the height and width of the video frames,  while \(T\)\ refers to the total number of frames.

The specific workflow of VWI is shown in the left part of Fig. \ref{fig_model}, which is summarized as follows: Initially, we obtain the initial features \({F}^\text{R}\)\ with redundant information through an encoder, which reduces frames annotated with wires to low-resolution features. Subsequently, the initial features \({F}^\text{R}\)\ are propagated to our proposed Raformer for eliminating redundant information and reconstructing the missing parts by employing an eight-layer Raformer. Finally, a decoder enlarges the obtained non-redundant features from Raformer and restores them back into the regenerated video \textbf{Y}. 
As depicted in the right part of Fig. \ref{fig_model}, our Raformer model is fundamentally composed of two distinct modules: the RAA Module and the SFA Module. The RAA module, forming the core of Raformer, is delicately designed to selectively sieve out non-crucial features at a coarse granularity, thereby optimizing the feature representation. The features then undergo further refinement and alignment through the SFA Module, which plays a pivotal role in ensuring the precise alignment of features.
The dashed lines in the figure illustrate the visualization process of removing redundancy in the VWI task. After dividing the input video frames into blocks, some blocks are redundant, and others are not. Our Raformer is capable of achieving redundant elimination.

In the following subsections, we initially introduce our Raformer framework in Section \ref{sec:Raformer}. Subsequently, the proposed RAA module and the SFA module within the Raformer are detailed in Sections \ref{sec:RAA} and \ref{sec:SFA}, respectively.
\subsection{Redundancy-Aware Transformer}\label{sec:Raformer}
A standard Raformer encompasses a Transformer Module, an RAA Module, and an SFA Module. Given a set of tokens \({F_{n-1}}\in \mathbb{R}^{T \times \left(\frac{H}{4}\right) \times \left(\frac{W}{4}\right) \times C}\) from the \({n-1}\) layer, we employ  mask-guided sparse video transformer\cite{Pronpainter_2023_ICCV} as the baseline, where the tokens are initially passed through the transformer module:
\begin{align}
&F_n^{'} = \text{MSA}(\mathrm{LN_1}(F_{n-1}) + F_{n-1}), \\
&F_n^{*} = \text{FFN}(\mathrm{LN_2}(F_n^{'})) + F_n^{'} .
\end{align}
After processing through the transformer, we obtain the features \(F_n^{*}\), which encompass both redundant and non-redundant information. To remove these redundancies and refine the feature representation,   \(F_n^{*}\) is subsequently fed into the RAA Module, resulting in:
\begin{equation}
{W}^\text{NR} = \text{RAA}(F_n^{*}),
\end{equation}
where \({W}^\text{NR}\) denotes the non-redundant features after the RAA module.

In this phase, the RAA Module is critical in eliminating and enhancing the feature representations obtained from the transformer. It ensures that only the most pertinent and non-redundant features are carried forward, thereby reducing noise and improving the efficiency of the subsequent processes.
Following the RAA process, the output \(F_n^\text{NR}\) is then introduced to the SFA Module, where it undergoes further refinement to ensure optimal alignment of the features, preparing them for the final reconstruction phase. This alignment process is critical in maintaining the coherence and continuity of the video sequences, ensuring that the inpainted regions blend seamlessly with the surrounding areas.
\begin{equation}
{F}_n^\text{NR} = \text{SFA}({W}^\text{NR})\in \mathbb{R}^{T \times \left(\frac{H}{4}\right) \times \left(\frac{W}{4}\right) \times C}.\
\end{equation}

Finally, we employ two learnable parameters, \( \beta \) and \(\gamma\), to perform a weighted summation of the original parameters and the residuals. Consequently, the final result manifests as:
\begin{equation}
F_{n} = \beta \cdot F_n^\text{NR} + \gamma \cdot F_n^{*},
\end{equation}
where \( F_{n} \) represents the final feature representation after incorporating the influence of both the original and non-redundant features, with \( \beta \) and \(\gamma\) are the weights that determine the contribution of each component to the final result.
Through the systematic integration of the Transformer Module, RAA Module, and SFA Module, the Raformer offers a robust and efficient solution for video inpainting, promising improved results over existing methods.

\subsection{Redundancy-Aware Attention Module}\label{sec:RAA}
For VWI, a solution can be realized  through video inpainting models if \textbf{M} skilfully masks the wires in each frame. Considering the example shown in Fig. \ref{house_masks},  we observe that certain temporal and spatial segments gain significant importance during the video inpainting process, especially when confronted by masks with slender straight-line characteristics. However, within these demarcated segments, there are regions that have minimal impact and do not actively contribute to the inpainting process.  To address this problem, we deliberately design the RAA Module to reduce redundant information at a coarse-grained level. As shown in Fig. \ref{fig_model}, our RAA Module workflow is as follows.

Commencing the process, we generate a C-channel feature map 
\(
{F}^\text{R} = E\left[ ({X}_t \odot {M}_t)_{t=1}^{T} \right] \in \mathbb{R}^{ T \times \left(\frac{H}{4}\right) \times \left(\frac{W}{4}\right) \times C}
\) through the encoder \(E\), which inherently contains both redundant and non-redundant information.
Our objective is to devise a mechanism at a coarse granularity level to efficiently discern and isolate the non-redundant features, thereby eliminating less pertinent data.
To achieve this, the feature map \({F}^\text{R}\) generated by the encoder, undergoes additional refinement through a transformer layer, resulting in the enhanced feature map \({F}^{*}\). Taking the \( t \)-th frame as an example, this refined frame is represented as \({F}_t^{*}\). To illustrate the coarse granularity removal of redundant information, consider the spatial division of these features into windows. Each window, sized \(h\) by \(w\), is designated as \({W}_{ij}^{t}\), with \(i\) and \(j\) indicating its row and column positions. The formulation of this process is as follows:
\begin{equation}
\begin{aligned}
{W}^{t} &= \sum_{ij}\{{W}_{ij}^{t}\} = \sum_{ij}\{{W}_{ij}^\text{R}, {W}_{ij}^\text{NR}\},
\end{aligned}
\end{equation}
where ${W}^{t} = \text{Partition}({(F}_t^{*}, h, w) \in \mathbb{R}^{n\times (h \times w) \times C}.$

Here, the partition segments ${F}_t$ into distinct windows by utilizing the spatial indices \( i \) and \( j \) for demarcating feature subsets, which include redundant (${W}_{ij}^\text{R}$) and non-redundant (${W}_{ij}^\text{NR}$). Note that the partition is based on a defined window size encompassed by the parameters \( h \) and \( w \). 
After obtaining these windows, we computed the mean value across each window's \(h \times w\)  features with the expectation of retaining non-redundant windows while discarding the redundant ones at a coarse granularity level. Subsequently, the coarse-granularity features are fed into an attention block to derive attention scores. 

The process is detailed as follows.
For each window \({W}_{ij}^{t} \), we average all the feature values to generate a new set of features. This step is represented with the following formula:
\begin{equation}
{\bar{W}}_{ij}^t = \frac{1}{h \times w} \sum_{p=1}^{h}\sum_{q=1}^{w} {W}_{ij}^{t}(p, q),
\end{equation}
where \({\bar{W}}_{ij}^t \) represents the set of coarse-granularity features, which are fed into a self-attention block post-acquisition to ascertain their individual levels of redundancy.
\begin{equation}
\begin{aligned}
\text{AW} = \text{Softmax}({Q} {K}^T), \\
\end{aligned}
\end{equation}
where ${Q}, {K} = \mathrm{LN_Q} ({\bar{W}}_{ij}^t) , \,  \mathrm{LN_k} ({\bar{W}}_{ij}^t).$ LN represents Layer Normalization for stabilizing feature inputs, and AW signifies Attention Weights highlighting key data features.

However, determining the redundancy level of each window based solely on self-attention scores derived from its individual frame \( t \) is not advisable, as it neglects the vital temporal information present across different frames. Hence, it is essential to extend our approach to include both temporal and spatial dimensions, enabling the elimination of redundant information across all frames. To this end, we introduce a formulation that leverages information across a series of frames, enhancing the robustness and efficiency of the video restoration process. To incorporate the temporal dimension, we redefine ${Q}$ and ${K}$ to span across all frames from 1 to \(T\), as:
\begin{align}
\label{eq:Q}
{Q} &= [\mathrm{LN_Q} ({\bar{W}}_{ij}^{1}), \mathrm{LN_Q} ({\bar{W}}_{ij}^{2}), \ldots, \mathrm{LN_Q} ({\bar{W}}_{ij}^{T})], \\
\label{eq:K}
{K} &= [\mathrm{LN_k} ({\bar{W}}_{ij}^{1}), \mathrm{LN_K} ({\bar{W}}_{ij}^{2}), \ldots, \mathrm{LN_k} ({\bar{W}}_{ij}^{T})].
\end{align}

Thus, the attention weights explicitly describe the relative importance of each feature vector in the windowed feature set in both temporal and spatial dimensions. Following this, a two-stage process is implemented to identify and preserve the least redundant information in our feature set. Firstly, the importance of each window is determined by summing over the second dimension of the attention weights. Secondly, the top \(k\) windows are selected based on these calculated importance scores. We depict this process concisely in the equation:
\begin{equation}
{W}^\text{NR} = \text{Top}_k \left( \sum_{d=2} \text{AW} \right)\in \mathbb{R}^{T\times k\times (h \times w) \times C}.
\end{equation}

In this manner, we concentrate on the most informative sections, eliminating redundant windows to retain only the crucial, non-redundant features, thereby refining the feature set to emphasize the most significant portions of the windowed feature set. It is worth noting that we empirically chose \( k \) to be half of the total number of windows, which will be validated in subsequent ablation studies. 

\subsection{Soft Feature Alignment Module}\label{sec:SFA}
Leveraging the refined features obtained through the RAA module, the SFA Module assumes a crucial role in aligning these feature representations, facilitating enhanced outcomes in video inpainting. The incorporation of the SFA module is designed to improve compatibility and adaptability across a variety of transformer architectures, paving the way for a universally adaptable framework. This initiative is central to fostering a ``plug-and-play'' environment, elevating the versatility of our model. 
To elucidate the SFA module further, upon securing the non-redundant windows \(W\), our first step is to restore these windows to their initial feature expression form, defined as:
\begin{equation}
\label{eq:alpha}
F^\alpha = \text{Reverse}(W^\text{NR}) \in \mathbb{R}^{4\alpha \times T \times \left(\frac{H}{8}\right) \times \left(\frac{W}{8}\right) \times C}.
\end{equation}

In this context, \(\alpha\) is governed by the relation \(\alpha = \frac{k}{n}\), where \(n\) denotes the initial number of windows crafted from the features, and \(k\) stands for the count of non-redundant windows selected for retention. Our empirical analysis illustrates that setting \(\alpha\) to 0.5 yields optimal results, a finding that will be further substantiated in the subsequent experimental. 
Nonetheless, when the value of \(\alpha\) diminishes below 0.25, a dimension duplication strategy is implemented to sustain a minimum threshold of 1, thereby securing the essential characteristics of the feature space. 
To better align the features, we enlist the aid of the Soft Split (SS) and Soft Composite (SC) techniques\cite{fuseformer}. In the employed scheme, the initial feature representation, denoted as \( F \), first undergoes a transformation facilitated by the SC module, retaining its original dimensionality without any reduction. Subsequently, the 4\(\alpha\) dimensional space is funneled through a two-phase feature transformation process, which integrates it back into its original feature dimensions, thereby reducing the dimensionality to its initial state.  This critical process ensures the restoration of \( F_{}^{NR} \) to its primal dimensional size, enabling the preservation of essential characteristics while availing it for further analytical operations. To foster better integration of features, we employ the SS technique after the features have passed through the SC and a two-phase feature transformation process, enhancing its readiness to be input into subsequent transformer modules. This orchestrated pathway, leveraging SC, SS, and a two-phase feature transformation process, fosters a meticulous transformation and reinstatement of the original dimensional scope, thereby establishing a robust framework for feature representation. Thus, the SFA module is expressed as the following equation:
  \begin{equation}
   F^{\alpha'} = \text{Conv2d}\left(\text{Upsampling}\left(\text{SC}\left(F^{\alpha}\right)\right)\right),
   \end{equation}
where Conv2d refers to a convolution operation with a 3x3 convolutional kernel and a stride of 1, applied to the upsampled SC features \( F^{\alpha} \). The final transformation is represented by the following equation:
\begin{equation}
F_{}^\text{NR} = \text{SS}\left(\text{Conv2d}\left(\text{LRelu}\left(F^{\alpha'}\right)\right)\right) \in \mathbb{R}^{ T \times \left(\frac{H}{4}\right) \times \left(\frac{W}{4}\right) \times C},
\end{equation}
where the SS technique is applied to the Convolved and LRelu-activated features \( F^{\alpha'} \), resulting in \( F_{}^{NR} \), which represents the non-redundant features. After being aligned with the initial input features, these non-redundant features are now ready for further processing in the Raformer framework.

\subsection{Training Objectives}
We employ two loss functions to train our Raformer: L1 loss and adversarial loss. The L1 captures the pixel-level discrepancies between the output image \( {Y} \) and the ground truth \( X \). Given its objective nature, it offers a straightforward measurement of the reconstruction quality, which is formulated as:
\begin{equation}
\mathcal{L}_{\text{rec}} = \frac{\sum (Y - X) \odot M}{\sum |M|} + \frac{\sum (Y - X) \odot (1 - M)}{\sum |1 - M|}.
\end{equation}

In addition, we leverage a T-PatchGAN\cite{Free-form} based discriminator to further enhance the reconstruction quality, encouraging the network to generate results that are more coherent and visually pleasing. The adversarial loss for the discriminator \( D \) is defined as:
\begin{equation}
\mathcal{L}_{D} = \mathbb{E}_{x}[\text{ReLU}(1 - D(G(x)))] + \mathbb{E}_{x}[\text{ReLU}(1 + D(x))].
\end{equation}

The adversarial loss for Raformer is formulated as:
\begin{equation}
\mathcal{L}_{adv} = -\mathbb{E}_{x}[D(G(x))] .
\end{equation}

Finally, the overall objective function for training the Raformer is formalized as follows:
\begin{equation}
\mathcal{L}_{\text{total}} = \mathcal{L}_{\text{rec}} +  \lambda_{\text{adv}} \cdot \mathcal{L}_{\text{adv}} ,
\end{equation}
where the adversarial weight \(\lambda_{\text{adv}}\) is empirically set to 0.01 as in \cite{E2FGVI,Pronpainter_2023_ICCV}.

\section{Experiments}
\begin{table*}[!ht]
\centering
\caption{Quantitative Results of Video Inpainting on WRV2 and DAVIS\cite{DAVIS} Dataset. Note That the Results of All the Competing Methods Are Re-Implemented by Running Their Released Codes with No Change. The Best and Second-Best Results Are Highlighted in Bold and Underlined.}
\renewcommand\arraystretch{1.1}

\label{tab:quantitative}
\begin{tabular*}{0.9\textwidth}{@{\extracolsep{\fill}}lccccccccc@{}}
\bottomrule 
& \multicolumn{5}{c}{\textbf{WRV2}} & \multicolumn{4}{c}{\textbf{DAVIS}} \\
\cline{2-6} \cline{7-10}  \\ [-2ex]
 \textbf{Models} & \textbf{PSNR $\uparrow$} & \textbf{PSNR* $\uparrow$} & \textbf{SSIM $\uparrow$} & \textbf{VFID $\downarrow$} & \textbf{LPIPS $\downarrow$}  & \textbf{PSNR $\uparrow$} & \textbf{SSIM $\uparrow$} & \textbf{VFID $\downarrow$} & \textbf{LPIPS $\downarrow$} \\ 
\bottomrule 
VINet\cite{kim2019deep} & 38.26 &32.75 & 0.9724 & 0.311 & 0.0473 & 28.03  & 0.9312 & 0.273 & 0.0656 \\
LGTSM\cite{chang2019learnable} & 38.47 &32.98 & 0.9713 & 0.307 & 0.0466 & 28.27 & 0.9349 & 0.284 & 0.0623 \\
FGVC\cite{Gao-ECCV-FGVC} & 39.55 &34.03 & 0.9818 & 0.255 & 0.0425 & 29.56 & 0.9566 & 0.266 & 0.0544 \\
STTN\cite{STTN} & 39.72 &34.31 & 0.9768 & 0.243 & 0.0393 & 29.41 & 0.9505 & 0.243 & 0.0537  \\ 
FuseFormer\cite{fuseformer} & 40.46 &34.98 & 0.9896 & 0.229 & 0.0355 & 29.76 & 0.9650 & 0.210 & 0.0456 \\
G2LP\cite{ji2022g2lp} & 40.50 &35.07 & 0.9878 & 0.182 & 0.0295 & 30.23 & 0.9664 & 0.179 & 0.0450 \\
E2FGVI\cite{E2FGVI} & 40.84&35.44 & \underline{0.9903} & 0.110 & 0.0255 & 30.72 & 0.9685 & 0.164 & 0.0449 \\
ProPainter\cite{Pronpainter_2023_ICCV} & \underline{41.16} &\underline{35.70}& 0.9889 & \underline{0.094} & \underline{0.0209} & \underline{31.48} & \underline{0.9739} & \underline{0.133} &\textbf{0.0302} \\
Ours & \textbf{41.70} & \textbf{36.20} & \textbf{0.9909} & \textbf{0.088} & \textbf{0.0180} & \textbf{31.98} & \textbf{0.9750} & \textbf{0.125} & \underline{0.0332} \\
\bottomrule 
\end{tabular*}
\end{table*}
\subsection{Datasets and Evaluation Metrics}
Besides our WRV2 dataset, we also select the popular DAVIS dataset \cite{DAVIS} to test the effectiveness of our model on traditional video inpainting.
Following the previous research \cite{E2FGVI,Pronpainter_2023_ICCV}, we partition the DAVIS dataset into two segments for our experiments: 100 video sets are utilized for training, and 50 video sets are reserved for testing. 
Notably, for the testing videos in the DAVIS dataset, we employ the same PP masks in E2FGVI\cite{E2FGVI}, and ProPainter\cite{Pronpainter_2023_ICCV} for a fair and consistent comparison.

For evaluating the quality of our inpainting results, we employ the popular PSNR and SSIM\cite{wang2004image} as the evaluation metrics.  
When evaluating the WRV2 dataset, we also calculate the average of the PSNR within the smallest bounding rectangles of the areas requiring wire removal, denoted as PSNR*.  
By cropping local windows and conducting PSNR calculations within them, PSNR* ensures evaluations focus on critical restoration areas, offering a more accurate measure of repair quality.
Furthermore, we apply VFID\cite{lyu2019advances} to assess visual fidelity and LPIPS\cite{zhang2018perceptual} for quantifying perceptual differences, offering a comprehensive evaluation of video inpainting quality by addressing both interpolation distortion and adherence to human visual standards.
In our evaluation, higher PSNR and PSNR* values denote superior image restoration and specific inpainting accuracy, respectively, while higher SSIM signifies greater fidelity to the original. Conversely, lower VFID and LPIPS scores highlight minimal distortion and closer alignment with human visual perception.
\subsection{Implementation Details}
In our research, we employ a consistent training framework for all models across both the WRV2 and DAVIS\cite{DAVIS} datasets. This framework involves a uniform learning rate of 1e-4, a batch size 4, and training conducted on two RTX 4090 GPUs.
Additionally, we standardize the configuration for all transformer-based models to have a local video sequence length of 5, in line with that of E2FGVI\cite{E2FGVI} and FuseFormer\cite{fuseformer}, ensuring a consistent baseline for comparison across all methods. Furthermore, for consistency in our experiments, all videos are resized to 432 × 240 resolution for training and evaluation.

For the WRV2 dataset, the training of all models is conducted for 350K iterations with PWS masks. Meanwhile, for the DAVIS dataset, which includes only 100 training videos, the training duration is adjusted to 140K iterations. For a fair comparison mirroring previous research\cite{STTN,fuseformer,E2FGVI,ji2022g2lp,Pronpainter_2023_ICCV}, models trained on the DAVIS dataset use the PP masks algorithm for generating masks.

\subsection{Comparisons with State-of-the-Art Approaches}
We quantitatively compare our Raformer method against various state-of-the-art models, including VINet\cite{kim2019deep}, LGTSM\cite{chang2019learnable}, FGVC\cite{Gao-ECCV-FGVC}, STTN\cite{STTN}, FuseFormer\cite{fuseformer}, G2LP\cite{ji2022g2lp}, E2FGVI\cite{E2FGVI} and ProPainter\cite{Pronpainter_2023_ICCV}. All the results are shown in Table \ref{tab:quantitative}.

\subsubsection{\textbf{Results on the WRV2 Dataset}}
For the WRV2 dataset, we focus on the challenging task of wire removal in videos. Our Raformer model sets a new benchmark in video inpainting tasks, leading the field across all key metrics. 

It achieves a groundbreaking PSNR of 41.70 and a PSNR* of 36.20, surpassing the second-best model by approximately 0.54 in PSNR and by 0.50 in PSNR*. Additionally, it records an SSIM of 0.9909, the highest among the compared models. This notable performance in PSNR and PSNR* underscores superior capability in restoring high-quality images, particularly in the focused assessment of wire removal areas. The relatively minor gap in SSIM stems from the specific characteristics of the wire removal task, where masks are usually small and elongated, rendering the perceptual differences less noticeable.
Additionally, our Raformer excels in video fidelity and perceptual similarity, as indicated by its leading scores in VFID and LPIPS, both the lowest among all models evaluated. Our method demonstrates its effectiveness in producing videos that are not only visually coherent but also closely aligned with human visual perception. 

\begin{table}[!t]
    \centering
    \caption{Wire Inpainting Performance Across Different Training Sets}
    \label{tab:improved_layout_merged}
    \begin{tabular}{c c S[table-format=2.2] c S[table-format=1.4]}
    \toprule
    {Testing Set} & {Training Set} & {{PSNR $\uparrow$}} & {{PSNR* $\uparrow$}} & {{SSIM $\uparrow$}} \\
    \midrule
    \multirow{3}{*}{WRV} & DAVIS\cite{DAVIS} & 42.83 & 37.34 & 0.9918 \\
                         & WRV\cite{ji2022g2lp} & 43.16 & 37.68 & 0.9939 \\
                         & WRV2 & \textbf{44.42} & \textbf{38.92} & \textbf{0.9949} \\
    \addlinespace
    \multirow{3}{*}{WRV2} & DAVIS\cite{DAVIS} & 38.32 & 32.82 & 0.9831 \\
                          & WRV\cite{ji2022g2lp} & 38.57 & 33.05 & 0.9865 \\
                          & WRV2 & \textbf{41.14} & \textbf{35.72} & \textbf{0.9910} \\
    \bottomrule
    \end{tabular}
\end{table}

\subsubsection{\textbf{Results on DAVIS Dataset}}
Turning to the DAVIS dataset, renowned for its diverse and complex video inpainting scenarios, our Raformer method again sets new benchmarks. 

On this dataset, our model maintains its lead with the highest PSNR score of  31.98 and the top SSIM score of 0.9750. These scores are significantly higher than the next best model, ProPainter, which achieves a PSNR of 31.48 and an SSIM of 0.9739. 

Transitioning to the metric of VFID, our Raformer model again outperforms the competition with the lowest score of 0.125, indicating superior video fidelity. However, it is important to note a different outcome in the LPIPS metric. Although our model achieves a commendable LPIPS score of 0.0332, it ranks second to ProPainter, which leads with an LPIPS score of 0.0302. 
The DAVIS dataset employs PP masks for both training and testing purposes. This approach contrasts markedly with the PWS masks that our model is specifically optimized to handle within the context of VWI.
These masks often cover substantial regions with backgrounds, presenting a distinct type of inpainting challenge. Our model, being tailored for VWI, is optimized for thin, linear masks. This specialized focus might lead to slightly less alignment with human visual perception in terms of textural details on the DAVIS dataset, as reflected in the LPIPS score. Nonetheless, our model still achieves the best performance in the other three metrics on the DAVIS dataset, demonstrating its overall effectiveness in video inpainting tasks. The outstanding performance on the DAVIS dataset demonstrates that although our Raformer is specifically designed for the VWI task, it is equally effective in general video inpainting. This efficacy is due to the RAA and SFA modules in our Raformer model, which are crafted to reduce unnecessary data and focus the model on the most beneficial aspects for inpainting, enhancing performance across various scenarios.

Overall, the quantitative evaluation on both WRV2 and DAVIS datasets firmly establishes the superiority of our Raformer method in video inpainting, especially in VWI.

\subsection{Ablation Studies}
In this section, we conduct extensive ablation studies, including the validity of our proposed WRV2 dataset, the effectiveness of each component of Raformer, the impact of \(\alpha\), and the influence of different mask types. 
To maintain uniformity and objectivity, each model is trained on the WRV2 dataset with the same training details.
Notably, except for the study examining the RAA Module with Parameter \(\alpha\), we consistently set \(\alpha\) to 1/2 across all experiments. 

\begin{table}[!t]
    \centering
    \caption{Ablation Study for Different Module Components, Symbol * Represents Non-Temporal-Aware RAA on the WRV2 Dataset.}
    \label{tab:ablation_study}
    \begin{tabular}{cccccc}
    \toprule
        Model&RAA & SFA & PSNR $\uparrow$ & PSNR* $\uparrow$ & SSIM $\uparrow$ \\
    \midrule
        1&- & - & 41.16  & 35.70 & 0.9889 \\
        2&* & - & 41.21  & 35.77 & 0.9890 \\
        3&$\checkmark$ & - & 41.52  & 35.99 & 0.9893 \\
        4&$\checkmark$ & $\checkmark$ & \textbf{41.70}  & \textbf{36.20} & \textbf{0.9909} \\
    \bottomrule
    \end{tabular}
\end{table}

\subsubsection{\textbf{Dataset Validity Verification}}
To assess the effectiveness of WRV2 dataset in VWI, especially against the previous WRV\cite{ji2022g2lp} dataset, we train Raformer on DAVIS\cite{DAVIS}, WRV\cite{ji2022g2lp}, and WRV2, and then evaluate their performance on the WRV and WRV2 test sets, respectively.
For the WRV dataset, we randomly select 100 video sets for training and 50 sets for testing. 
Considering the test set of  WRV2 dataset encompasses all 150 videos from WRV, we specifically exclude the 100 videos in WRV employed for training when evaluating on WRV2.
Since the WRV dataset only has 100 videos for training, we conduct 140K iterations on each dataset, with all other parameter settings remaining identical. 
Results in Table \ref{tab:improved_layout_merged} show that training on WRV2 consistently yields the highest performance across both the WRV and WRV2 test sets.
This comparison underscores the ability of WRV2 to present a more challenging and comprehensive benchmark for VWI tasks.

\subsubsection{\textbf{Effectiveness of RAA and SFA modules}} To validate the individual contributions of the RAA and SFA modules, we conduct comprehensive ablation studies, as shown in Table \ref{tab:ablation_study}. Specifically, Model 1 serves as the baseline, without RAA and SFA modules. 
Model 2 implements this non-temporal-aware version of RAA by essentially omitting the temporal joint from Eq. (\ref{eq:Q}) and Eq. (\ref{eq:K}). This modification allows us to assess the impact of excluding the temporal component from the RAA module. Despite the lack of temporal awareness, improvements in Model 2 indicate that even a simplified RAA can positively influence video inpainting quality.
\begin{figure}[!ht]
\centering
\includegraphics[width=0.45\textwidth]{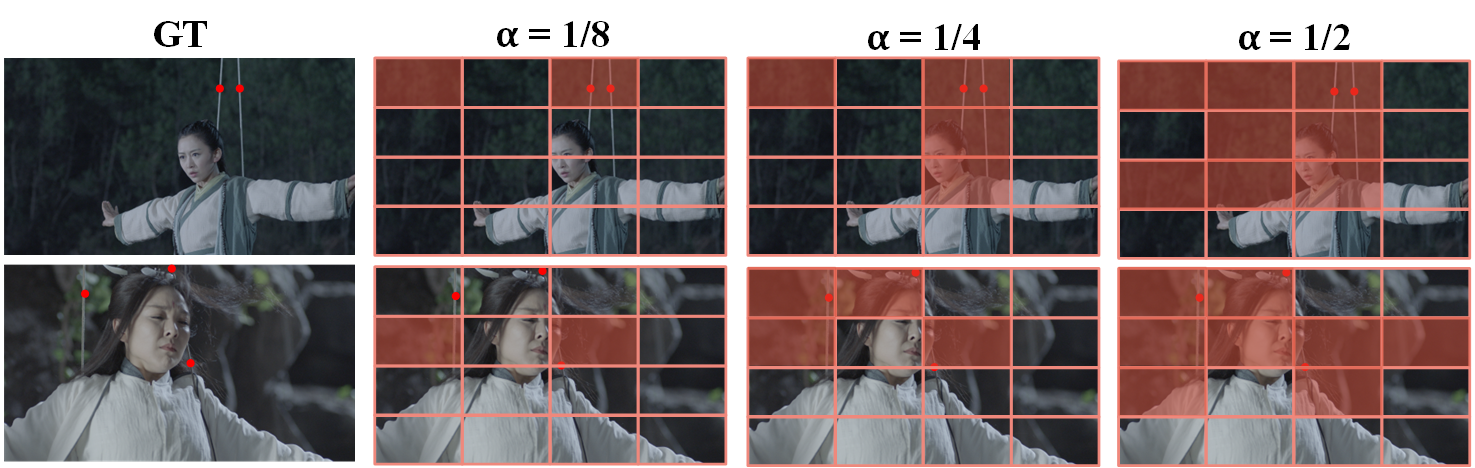}
\caption{Comparison of the effects of different \(\alpha\) values on determining focus areas: The original frames (GT) are shown on the left, with subsequent frames displaying expanded attention areas for \(\alpha\) values of 1/8, 1/4, and 1/2, as indicated by the red grids overlaying the images. The areas highlighted with red points identify the wires that need to be inpainted. Note that the visualization results are derived from the attention weights extracted from the fourth layer out of the eight layers in our Raformer.}
\label{fig_abtion}
\end{figure}
Integrating the complete RAA module (indicated by $\checkmark$) in Model 3 leads to significant enhancements, confirming its effectiveness in emphasizing salient features and minimizing redundancy in the VWI process.
It is important to note that for Models 2 and 3, we employ simple resizing to align features for RAA, maintaining focus on the RAA module's impact without complicating factors from advanced alignment methods.

Incorporating the SFA module alongside RAA in Model 4 culminates in the highest performance across all metrics. The integration of SFA, which excels in aligning activated features with high precision, is evidently pivotal. 
In summary, our ablation study clearly illustrates the integral roles of both RAA and SFA modules in attaining top-tier results in VWI. 
\begin{figure*}[!t]
\centering
\includegraphics[width=0.9\textwidth]{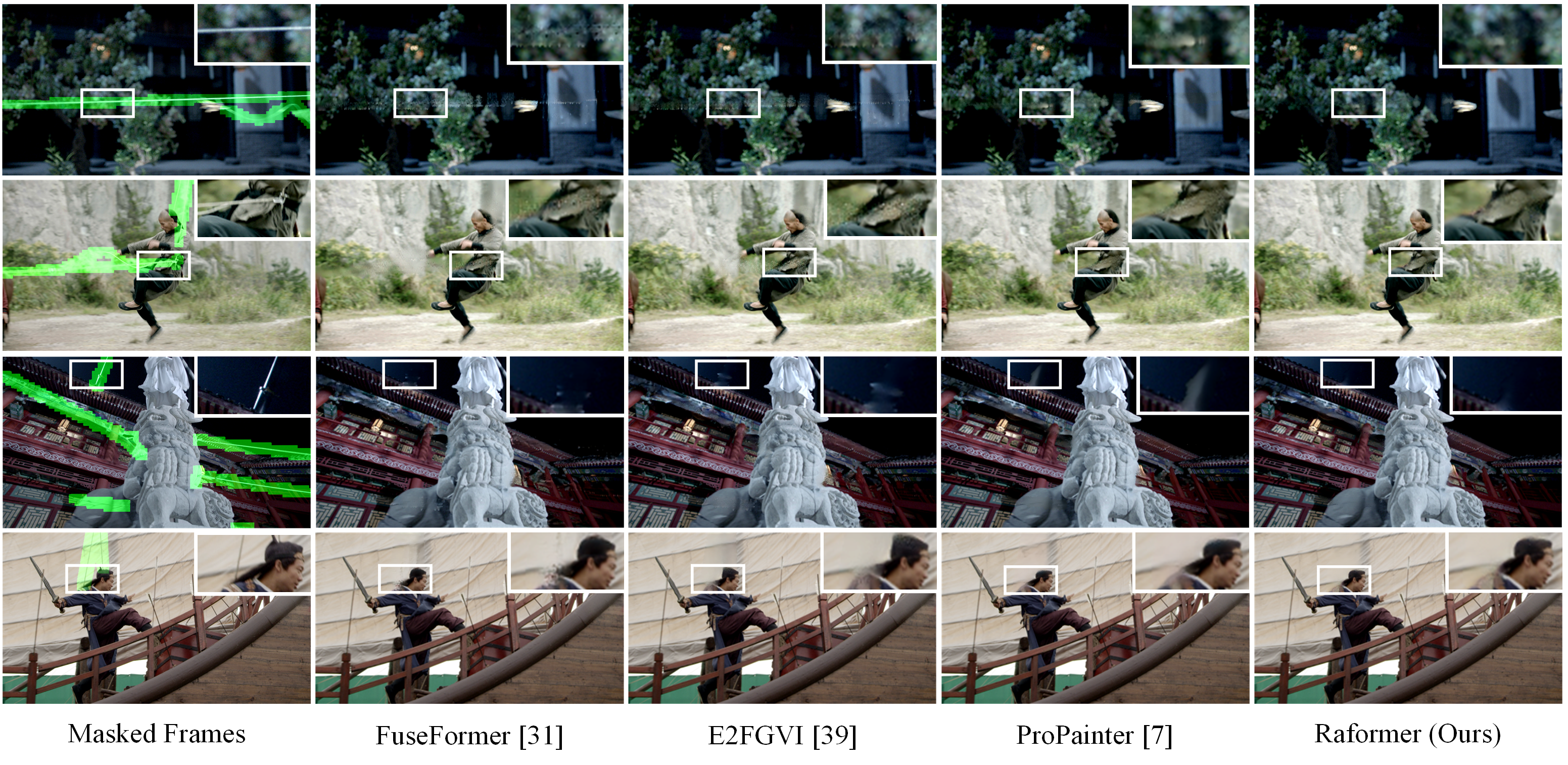}
\caption{Qualitative comparisons with FuseFormer\cite{fuseformer}, E2FGVI\cite{E2FGVI}, and ProPainter\cite{Pronpainter_2023_ICCV} on WRV2 dataset. In these comparisons, small white boxes within the images demarcate the specific areas targeted for inpainting, primarily the wire sections intended for removal. Additionally, the large white box located in the top right corner of each image serves as an enlarged, proportional view of these small white boxes, thereby drawing focus to the crucial details.}
\label{fig_reslut}
\end{figure*}
\subsubsection{\textbf{Effectiveness of the Parameter} \(\alpha\)} We then delve into the impact of the parameter \(\alpha\) in Eq. (\ref{eq:alpha}). This parameter is pivotal as it dictates the proportion of non-redundant blocks selected from the total number of blocks.
We experiment with different values of \(\alpha\) and document the results in our evaluation metric. 
The findings in Table \ref{tab:ablation_alpha} indicate that an \(\alpha\) value of 0.5 is optimal, with lower values like 1/8 and 1/4 reducing performance, underscoring the importance of choosing the right proportion of informative blocks for effective inpainting.

Additionally, we present attention score visualizations in Fig. \ref{fig_abtion}, offering a qualitative perspective that aligns with our quantitative findings. 
These visualizations highlight the model's attention distribution, confirming the benefits of the optimal \(\alpha\) setting. 
Our analysis reveals that an \(\alpha\) value of 1/2 is particularly effective for VWI. When \(\alpha\) is set to lower values like 1/8 or 1/4, the model's focus area is too narrow, insufficiently addressing the wire regions that require inpainting. 
On the other hand, an \(\alpha\) setting of 1/2 appropriately expands the attention area, ensuring comprehensive coverage of the wire regions in need of inpainting. It is worth noting that the windows selected by RAA do not always align with human intuition. For instance, at an \(\alpha\) = 1/8, RAA does not prioritize windows with wires but instead selects both windows with and without wires. This phenomenon reflects the nature of video inpainting, which relies on referencing information from other frames to fill missing areas in the current frame. Hence, RAA focuses on identifying non-redundant patches that are essential for inpainting across frames. 

\begin{table}[!t]
\centering
\caption{Impacts of Parameter $\alpha$ on the WRV2 Dataset}
\renewcommand\arraystretch{1.1}
\label{tab:ablation_alpha}
\begin{tabular}{cccc}
\toprule
$\alpha$ &  PSNR $\uparrow$ & PSNR* $\uparrow$ & SSIM $\uparrow$  \\
\midrule
1/8 & 41.47 & 35.93 & 0.9895 \\
1/4 & 41.54 & 36.01 & 0.9899 \\
1/2 & \textbf{41.70} & \textbf{36.20} & \textbf{0.9909} \\
\bottomrule
\end{tabular}
\end{table}
\begin{table}[!t]
    \centering
    \caption{Impact of Different Mask Types on Inpainting Performance on WRV2 Dataset}
    \renewcommand\arraystretch{1.1}
    \label{tab:mask_type_impact}
    \begin{tabular}{cccc}
    \toprule
    \textbf{Mask Type} &  PSNR $\uparrow$ & PSNR* $\uparrow$ & SSIM $\uparrow$ \\
    \midrule
    PP Masks& 41.30& 35.74 & 0.9892\\
    PWS Masks (Ours) & \textbf{41.78}& \textbf{36.26} & \textbf{0.9911} \\
    \bottomrule
    \end{tabular}
    \end{table}
\subsubsection{\textbf{Effectiveness of  Different Mask Types}}\label{sec:PWS}
In our ablation study to assess the effectiveness of the PWS mask, we train Raformer on the WRV2 dataset with two different mask types: PP masks and our proposed PWS masks. 
As shown in Table \ref{tab:mask_type_impact}, the PWS masks outperform the PP masks in all metrics. The results underscore the effectiveness of our PWS masks in addressing the specific challenges of wire removal. By more accurately representing the slender characteristics of wires that are typically encountered in such tasks, our PWS masks are more aligned with the complexities present in actual wire removal scenarios in film production.
\subsection{Visualization and Analysis}
In our visualization analysis, all models are fairly trained on the WRV2 dataset for 350K iterations. The qualitative evaluation of our Raformer method, as illustrated in Fig. \ref{fig_reslut}, showcases its effectiveness in VWI. The comparative visuals, which include our method and three other recent approaches\cite{fuseformer,E2FGVI,Pronpainter_2023_ICCV}, reveal how all models tend to suffer from artifacts in VWI. However, it is observable that Raformer experiences the least amount of these artifacts. 
This improved performance is attributed to the implementation of the RAA and SFA modules in our framework, which selectively focus on essential details and redundant elimination.

Visually, Raformer demonstrates a superior capability in seamlessly blending the inpainted areas with the surrounding content, effectively making wire traces virtually undetectable. This leads to a more natural and coherent visual output compared to existing methods, which often exhibit noticeable artifacts in similar scenarios. For example, consider the scenario depicted in the first row of  Fig. \ref{fig_reslut}, where a wire is towing forward a cluster of feathers. The outcomes frequently include disjointed content marred by artifacts when employing alternative methods to remove the wire. In contrast, Raformer achieves a more authentic visual result, aligning more closely with human perception of reality. This enhanced performance is attributed to Raformer's ability to process features so that it identifies and discards redundancy. Consequently, this allows Raformer to leverage more relevant information from other frames, facilitating a more effective inpainting process.
Therefore, our method excels in maintaining the integrity and aesthetic continuity of the video scenes, a critical factor in professional film editing. 
\begin{figure}[!t]
\centering
\includegraphics[width=0.45\textwidth]{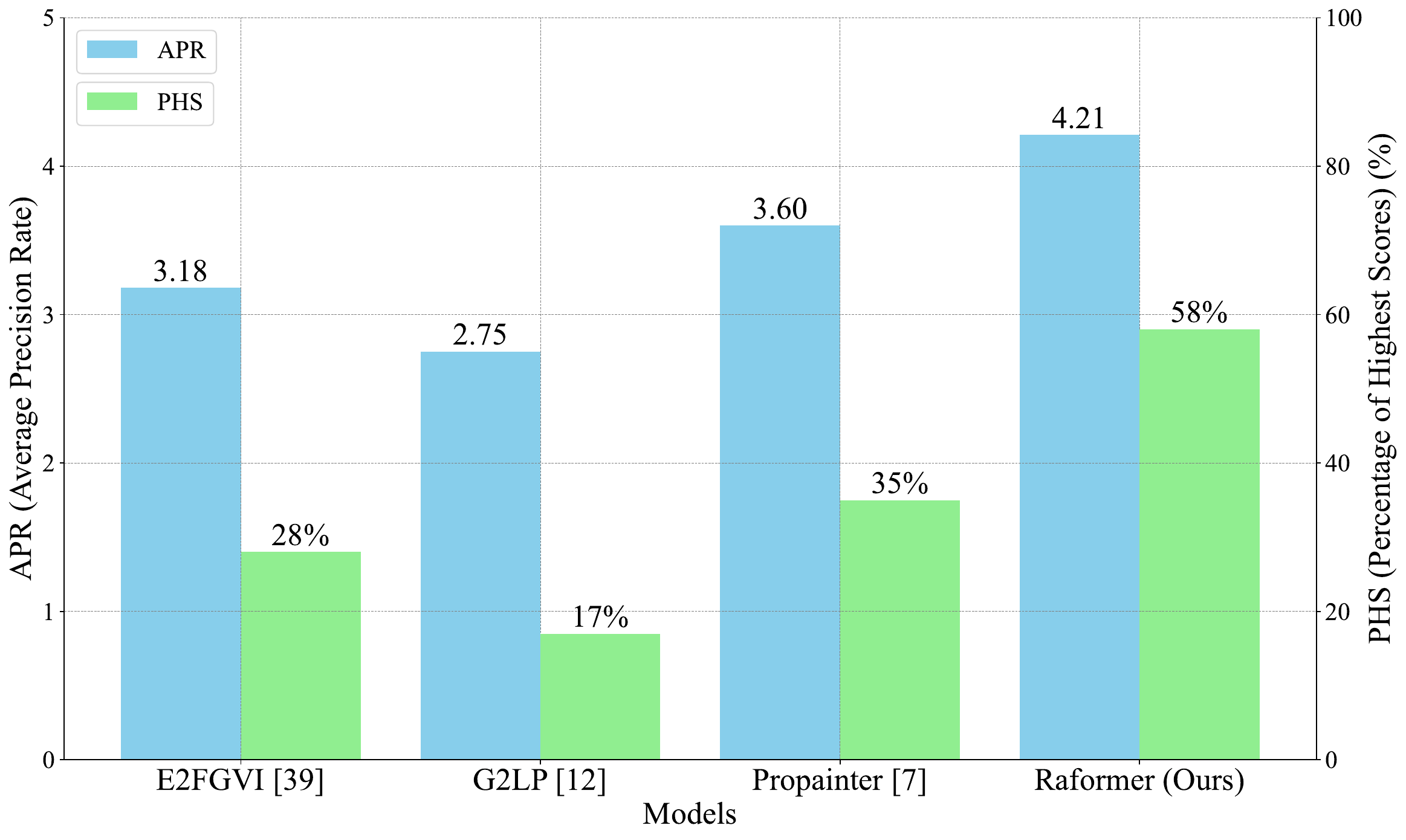}
\caption{Comparison of G2LP\cite{ji2022g2lp}, E2FGVI\cite{E2FGVI}, Propainter\cite{Pronpainter_2023_ICCV}, and our Raformer on both the APR and PHS metrics.}
\label{fig_usrstudy}
\end{figure}
\subsection{User Study}
We evaluate the performance of video processing models in human visual perception, including G2LP\cite{ji2022g2lp}, E2FGVI\cite{E2FGVI}, Propainter\cite{Pronpainter_2023_ICCV}, and our Raformer.
From the WRV2 dataset, we randomly select 12 video sets featuring authentic wire masks.
To gauge the effectiveness of our proposed model, we enlist 30 non-professional volunteers, aged 18 to 60, each of whom scores the videos on a 1 to 5 scale. This scale is chosen for its simplicity and ability to capture quality gradations effectively, with 1 indicating a failure and 5 signifying excellent execution without visible modifications. 

Volunteers are provided with detailed instructions on how to utilize this scoring system to ensure consistent and fair evaluation. They are afforded the flexibility to view the videos at their convenience, with options to adjust playback speed or examine individual frames closely, thus enabling a thorough and nuanced assessment. The outcomes of this study are quantified through two metrics: the Average Precision Rate (APR), representing the mean score across all evaluations, and the Percentage of Highest Scores (PHS), highlighting the proportion of videos that are rated highest by each method. As illustrated in Fig. \ref{fig_usrstudy}, our Raformer model achieves the best results in WRV2 dataset, indicating that users consider our approach to be more effective in accomplishing the VWI task.

\section{Conclusion}
This paper has introduced a new Wire Removal Video Dataset (WRV2) and a specialized mask generation algorithm named Pseudo Wire-Shaped Masks Generation to promote the development of advanced methods in the Video Wire Inpainting field. Then, we have presented a novel Redundancy-Aware Transformer model for VWI, in which the Redundancy-Aware Attention (RAA) module is designed to highlight essential content and eliminate redundant information, and the Soft Feature Alignment (SFA) module is introduced to align the features for seamless integration. Quantitative results reveal that our Raformer method establishes a new state-of-the-art (SOTA) on the WRV2 dataset and demonstrates impressive results on the traditional DAVIS dataset. Qualitative results further confirm that our Raformer method produces more coherent and integrated video outcomes. In the future, our focus will shift towards achieving a more efficient method aimed at significantly expediting the post-production process.
\bibliographystyle{ieeetr}
\bibliography{ref}

\newpage
\vfill

\end{document}